\documentclass[sigconf]{acmart}

\AtBeginDocument{%
  \providecommand\BibTeX{{%
    \normalfont B\kern-0.5em{\scshape i\kern-0.25em b}\kern-0.8em\TeX}}}

\setcopyright{rightsretained} 
\copyrightyear{2021} 
\acmYear{2021} 
\acmConference[CIKM '21]{Proceedings of the 30th ACM International Conference on Information and Knowledge Management}{November 1--5, 2021}{Virtual Event, QLD, Australia}
\acmBooktitle{Proceedings of the 30th ACM International Conference on Information and Knowledge Management (CIKM '21), November 1--5, 2021, Virtual Event, QLD, Australia}
\acmDOI{10.1145/3459637.3482111}
\acmISBN{978-1-4503-8446-9/21/11}

\usepackage{multirow}
\usepackage{graphicx}
\usepackage{amsmath}

\usepackage{bbm}

\usepackage{commath}

\usepackage{enumitem}
\usepackage{fixltx2e}

\usepackage[ruled,vlined]{algorithm2e}

\usepackage{romannum}

\usepackage{subcaption}
\usepackage{cleveref}

\begin{document}

\fancyhead{}

\title{GGP: A Graph-based Grouping Planner \\for Explicit Control of Long Text Generation}


\author{Xuming Lin}
\affiliation{%
  \institution{Alibaba Group}
  \streetaddress{}
  \city{}
  \country{}}
\email{xuming.lxm@alibaba-inc.com}

\author{Shaobo Cui}
\affiliation{%
  \institution{Alibaba Group}
  \streetaddress{}
  \city{}
  \country{}}
\email{yuanchun.csb@alibaba-inc.com}

\author{Zhongzhou Zhao}
\affiliation{%
  \institution{Alibaba Group}
  \streetaddress{}
  \city{}
  \country{}}
\email{zhongzhou.zhaozz@alibaba-inc.com}

\author{Wei Zhou}
\affiliation{%
  \institution{Alibaba Group}
  \streetaddress{}
  \city{}
  \country{}}
\email{fayi.zw@alibaba-inc.com}

\author{Ji Zhang}
\affiliation{%
  \institution{Alibaba Group}
  \streetaddress{}
  \city{}
  \country{}}
\email{zj122146@alibaba-inc.com}

\author{Haiqing Chen}
\affiliation{%
  \institution{Alibaba Group}
  \streetaddress{}
  \city{}
  \country{}}
\email{haiqing.chenhq@alibaba-inc.com}

\renewcommand{\shortauthors}{Xuming Lin, Shaobo Cui, Zhongzhou Zhao, Wei Zhou, Ji Zhang, Haiqing Chen}
\begin{abstract}
Existing data-driven methods can well handle short text generation. However, when applied to the long-text generation scenarios such as story generation or advertising text generation in the commercial scenario, these methods may generate illogical and uncontrollable texts. To address these aforementioned issues, we propose a graph-based grouping planner~(GGP) following the idea of first-plan-then-generate. Specifically, given a collection of key phrases, GGP firstly encodes these phrases into an instance-level sequential representation and a corpus-level graph-based representation separately. With these two synergic representations, we then regroup these phrases into a fine-grained plan, based on which we generate the final long text. We conduct our experiments on three long text generation datasets and the experimental results reveal that GGP significantly outperforms baselines, which proves that GGP can control the long text generation by knowing how to say and in what order.
\end{abstract}



\maketitle

\section{Introduction}
\begin{figure}[t]
  \centering
  \includegraphics[width=0.8\linewidth]{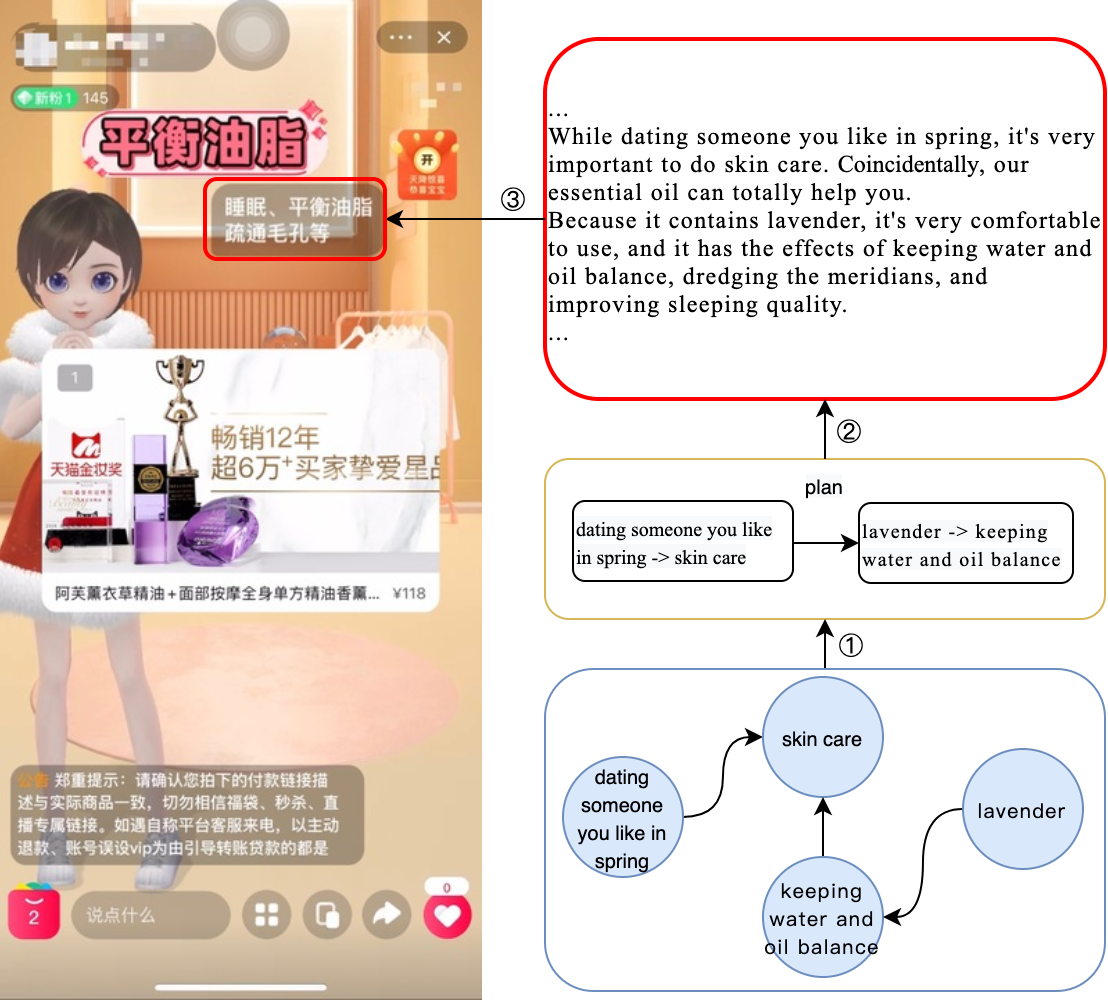}
  \caption{Application of our proposed GGP in our actual industrial scenario.}
  \label{fig:example}
\end{figure}

In recent years, live streaming is becoming an increasingly popular trend of sales in E-commerce. In live streaming, an anchor will attractively introduce the listed product items, and offer certain discounts or coupons, to facilitate user interaction and volume of transactions. 
However, script creation of product introduction is very time-consuming and requires professional sales experience. 
To alleviate this problem, we launched AliMe Avatar, an AI-powered Vtuber for automatically product broadcasting in the live-streaming sales scenario. Figure~\ref{fig:example} reveals how our virtual avatar broadcasts products with scripts created by our proposed planning-based methods. 
In AliMe Avatar, the process of script creation is abstracted into a first-plan-then-generate data-to-text task. The task of data-to-text generation is to generate a natural language description for given structured data~\cite{gatt2018survey}. Data-to-text methods have a wide range of applications in domains like automatic generation of weather forecasting, game report and production description. 
Table~\ref{tab:example} shows an example where the input data is a collection of key phrases, and the output texts are corresponding natural language descriptions generated from intermediate plans. 


Recently several published articles are inspired by the first-plan-then-generate pipeline~\cite{kukich1983design,mei2015talk} which includes content planning, sentence planning and surface realization. \citeauthor{moryossef2019step}~\cite{moryossef2019step} proposed a planning-based method to split the generation process into a symbolic text-planning stage followed by a neural generation stage. The text planner determines the information structure and expresses it unambiguously as a sequence of ordered trees from graph data. \citeauthor{puduppully2019data}~\cite{puduppully2019data} also decomposed the generation task into two stages with end-to-end methods instead. \citeauthor{zhao2020bridging}~\cite{zhao2020bridging} focused on planning from graph data and proposed a model named DualEnc. DualEnc is a dual encoding model that can not only incorporate the graph structure but can also cater to the linear structure of the output text. Considering the task of long text generation, \citeauthor{shao2019long}~\cite{shao2019long} proposed a planning-based hierarchical variational model~(PHVM) to generate long texts with plans.

However, these works ignore two points that are important in our scenarios: (1) Most of these works ignore the sentence planning process at the text planning stage and do not allow key phrase duplication. (2) The graph is extracted from the instance-level sample instead of the whole corpus, which is unable to learn regular patterns from a global perspective. Considering to overcome these shortcomings, we propose a GGP model. GGP combines sentence planning and text planning at the planning stage based on a corpus-level graph. Concretely, GGP can not only be aware of instance-level sequential representations but also corpus-level graph-based representations for each given key phrase. During the decoding process, GGP can regroup these encoded phrases into sentence-level plans and then paragraph-level plans in a fine-grained way. Finally, long texts are generated according to the generated plans. Moreover, recently \citeauthor{ribeiro2020investigating}~\cite{ribeiro2020investigating} and \citeauthor{kale2020text}~\cite{kale2020text} reveal that large scale pre-trained language models are evidenced by large improvements on data-to-text tasks. So in this situation, we conduct our experiments based on pre-trained language models to further prove the effectiveness of our proposed methods.




Our contributions are summarized as follows:
\begin{itemize}
\item We propose a grouping planner with a copy mechanism to first generate sentence-level plans and then paragraph-level plans in a fine-grained way with sentence planning and text planning together, in which key phrase duplication with any orders is allowed.

\item We combine the planner with graph neural networks to guide the plan generation with local instance-level information and global corpus-level knowledge learned from graphs.

\item We achieve state-of-the-art results on all planning related metrics on three public data-to-text datasets which are required to generate long texts, including the Advertising Text Generation(ATG) dataset from~\citeauthor{shao2019long}~\cite{kale2020text}, the Now You're Cooking dataset from~\citeauthor{kiddon2016globally}~\cite{kiddon2016globally} and the preprocessed Tao Describe~(TaoDesc) dataset from~\citeauthor{chen2019towards}~\cite{chen2019towards}.
\end{itemize}

\begin{table}[t]
  \caption{An example of structured data, plan and text from the ATG dataset. The plan is represented by a list of sequential groups.}
  \label{tab:example}
  \begin{tabular}{c p{7cm}}
  \hline 
    data & figure-flattering, aesthetic plaid, youthful, pure cotton, fresh, high-rise, irregular flounce, skirt \\
    \midrule
    plan & skirt, pure cotton; fresh, aesthetic plaid; figure-flattering, high-rise; irregular flounce, youthful \\
    \midrule
    text & This skirt is made of pure cotton, which is comfortable and breathable. The fresh and aesthetic plaid brings about lovely preppy style. The figure-flattering design, in particular, the high-rise design, heightens your waistline and perfects your body shape. The irregular flounce is youthful, making you look a bit more free and easy. \\
  \bottomrule
\end{tabular}
\end{table}



\section{Model}

\begin{figure*}
  \includegraphics[width=0.8\textwidth]{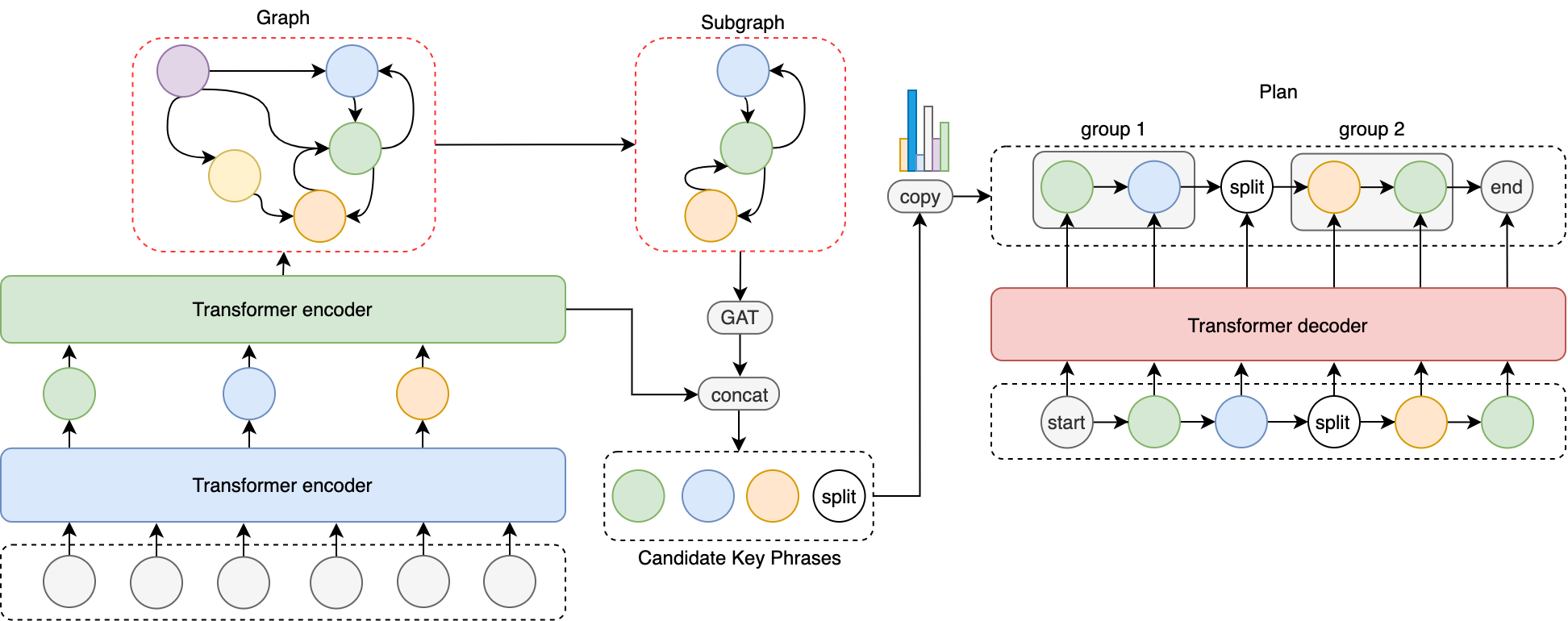}
  \caption{The model structure of GGP.}
  \Description{}
  \label{fig:model-struct}
\end{figure*}

\subsection{Model Overview}


Given a collection of key phrases $x=\{p_1,p_2,\dots,p_n\}$, ${p_i}$ represents a possible key phrase which consists of several tokens, in other words, ${p_i}=\{w^{i}_{1},\dots,w^{i}_{k}\}$. Our goal is to generate a context which can represent the content of the key phrase. Instead of the conventional straightforward \textit{first-plan-then-generate} approach, we adopt a graph-based grouping approach: (1) Encoding key phrases from the input into a combination of sequential representations and graph-based representations; (2) A plan $c=\{c_{11},\dots,c_{1m};$ $\dots;$ $c_{g1},\dots,c_{gl}\}$ is generated by picking and regrouping key phrases from ${x}$ instead of reordering these phrases only. For each ${c_{ij}}$, it means the key phrase is chosen as the ${jth}$ element of the ${ith}$ group. For example, in the ATG dataset, given a collection of key phrases $x$=\textit{\{figure-flattering, aesthetic plaid, youthful, pure cotton, fresh, high-rise, irregular flounce, skirt\}}, it's required to generate a plan according to this key phrase list. In this case, the plan can be generated like c=\{skirt, pure cotton; fresh, aesthetic plaid; figure-flattering, high-rise; irregular flounce, youthful\}, where $c_{11}$=skirt, and ';' means the group boundary. Finally the story is generated according to this plan.



\subsection{Hierarchical Sequential Encoding}
Given $n$ key phrases, a hierarchical transformer-based encoder~\cite{vaswani2017attention} is used to encode key phrases into a sequence of vector representations, and then these vectors are encoded again with another transformer to learn inter-relation among these key phrases. $w^{t}$ means the raw input tokens of each key phrase $p^{s}_{t}$. $c^{s}_{t}$ is the final key phrase representation after being encoded with the hierarchical sequential encoder.

\begin{equation}
    p^{s}_{t}=\text{Encoder}([w^{t}_{1:k};w^{1:n}])
\end{equation}

\begin{equation}
    c^{s}_{t}=\text{Encoder}([p^{s}_{t};p^{s}_{1:n}])
\end{equation}

\subsection{Graph Encoding}

To learn the graph-based representation, we first build a probability transition graph obtained from the statistics of the entire corpus, in which each node represents a key phrase. Then for each sample, we can build a subgraph according to the given key phrases. The nodes in this subgraph are encoded with a graph neural network. In our setting, we choose Graph Attention Networks~(GAT)~\cite{veli2018graph} instead of Graph Convolutional Networks~(GCN)~\cite{kipf2017semisupervised} as GAT performs slightly better in our experiments than GCN does. Let $M_g$ be the relation matrix of the current data input.

\begin{equation}
    c^{g}_{t}=\text{GAT}([M_g;c^{s}_{t}])
\end{equation}

\subsection{Grouping Copynet}
The graph-based representation may not preserve the original sequential information. In this way, we combine two representations from the graph encoder and the sequential encoder together. Specifically, given the graph-based representation $c^{g}_{t}$ and the sequential representation $c^{s}_{t}$, we merge these two vectors into one vector $m_{t}$ with MLPs:

\begin{equation}
    m_{t}=\text{MLP}([c^{g}_{t};c^{s}_{t}])
\end{equation}

Suppose there are $n$ key phrases in the key phrase list, a transformer-based decoder~\cite{vaswani2017attention} with a copy mechanism is used to pick key phrases from the input, and the generated plan is a sequence of groups. The white node in Figure~
\ref{fig:model-struct} represents the group boundary, and each group represents the plan of each sentence. For every decoding step, the decoder is required to discriminate whether it should be separated as a group boundary or pick a key phrase from the input collections. 

\begin{equation}
    z_{t}=\text{Decoder}([z_{t-1};y_{t-1};m_{1:n}])
\end{equation}

Finally, the loss function is calculated as cross entropy for each decoding step.


\section{Experiments}

\subsection{Dataset}

\begin{itemize}
\item{\textbf{Advertising Text Generation~(ATG)}}~\cite{shao2019long}: we used the same dataset from~\citeauthor{shao2019long}~\cite{shao2019long}, which consists of 119K pairs of Chinese advertising text.
\item{\textbf{Now You're Cooking~(Cooking)}}~\cite{kiddon2016globally}: we used the same dataset and pre-processing process from~\citeauthor{kiddon2016globally}~\cite{kiddon2016globally}. In the training set, the average recipe length is 102 tokens, and the vocabulary size of recipe text is 14,103.
\item{\textbf{Tao Describe~(TaoDesc)}}~\cite{chen2019towards}: TaoDesc dataset is from~\citeauthor{chen2019towards}~\cite{chen2019towards}. This dataset contains 2.1M pairs of product descriptions created by shop owners. We extract key phrases from these descriptions with our sequence labeling model. The format of this dataset is the same as the format of the ATG dataset.
\end{itemize}

\subsection{Baselines}
We compared our model with four strong baselines, including a pre-trained language model baseline, BART\cite{lewis2019bart}, and three planning-based baselines, PHVM, Step-By-Step and DualEnc.

\begin{itemize}
\item{\textbf{BART}}~\cite{lewis2019bart}: a pre-trained autoencoder whose input is a collection of key phrases or the generated plan, and output is the generated text in our settings. 

\item{\textbf{PHVM}}~\cite{shao2019long}: this model achieves state-of-the-art results on ATG. In our experiments, we use the exact implementation in \citeauthor{shao2019long}~\cite{shao2019long} as a strong baseline on ATG.

\item{\textbf{Random Planner~(RP)}}: it chooses the key phrases from the phrase list randomly as a plan.

\item{\textbf{Step-By-Step~(SBS)}}~\cite{moryossef2019step}: this method captures the division of facts into sentences and the ordering of the sentences. Again we use the exact implementation in ~\citeauthor{moryossef2019step}~\cite{moryossef2019step} on our datasets.

\item{\textbf{DualEnc~(DE)}}~\cite{zhao2020bridging}: this is a dual encoding model that can not only incorporate the graph structure but also can cater to the linear structure of the output text to extract plans. Our implementation is according to ~\citeauthor{zhao2020bridging}~\cite{zhao2020bridging} on our datasets.
\end{itemize}

\begin{table}[t]
  \centering
  \small
  \caption{Generation results on three test datasets evaluated by BLEU-4, PLAN BLEU-4 and PLAN ROUGE-L. We compare our methods with BART, PHVM, Step-By-Step, and DualEnc as our baselines. Our methods outperform all baselines on all metrics.}
  \label{tab:results}
\resizebox{0.485\textwidth}{!}{
  \begin{tabular}{p{1cm} p{1.5cm} p{1cm} p{1cm} p{1cm}}
    \toprule
    Dataset & Method & BLEU-4 & PB-4 & PR-L \\
    \midrule
     \multirow{5}{*}{ATG} & PHVM & 2.9 & 13.8 & 64.7 \\
     & BART & 4.0 & 17.5 & 66.3 \\
     & RP + BART & 2.6 & 5.9 & 52.8 \\
     & SBS + BART & 1.7 & 2.9 & 47.1 \\
     & DE + BART & 2.7 & 9.4 & 58.5 \\
     & GGP + BART & \textbf{4.3} & \textbf{20.8} & \textbf{68.7} \\
     \midrule
     \multirow{4}{*}{Cooking} & BART & 7.3 & 17.7 & 73.1 \\
     & RP + BART & 0.6 & 4.6 & 58.7 \\
     & SBS + BART & 0.5 & 4.4 & 45.1 \\
     & DE + BART & 1.4 & 12.0 & 55.7 \\
     & GGP + BART & \textbf{7.5} & \textbf{19.5} & \textbf{74.1} \\
     \midrule
     \multirow{4}{*}{TaoDesc} & BART & 14.6 & 13.7 & 59.4 \\
     & RP + BART & 13.0 & 5.6 & 51.8 \\
     & SBS + BART & 3.6 & 1.7 & 44.2 \\
     & DE + BART & 13.2 & 12.9 & 59.5 \\
     & GGP + BART & \textbf{16.8} & \textbf{16.8} & \textbf{61.6} \\
    \bottomrule
  \end{tabular}
  }
\end{table}

\subsection{Automatic Evaluation Metrics}
We adopted the following automatic metrics to evaluate the quality of generated outputs: (1) \textbf{BLEU-4}~\cite{papineni2002bleu}. (2)\textbf{PLAN BLEU-4~(PB-4)}~\cite{papineni2002bleu}: this metric is exactly BLEU-4, but hypotheses and references are generated plans and golden plans individually. (3) \textbf{PLAN ROUGE-L~(PR-L)}~\cite{lin2004rouge}: the same as PLAN BLEU-4, but the metric is ROUGE-L instead of BLEU-4. PLAN BLEU-4 and PLAN ROUGE-L measure the quality of generated plans while BLEU-4 focuses on measuring the generated text.

\subsection{Experimental Results}

\begin{table}[t]
  \centering
  \small
  \caption{Ablation study on the ATG dataset reveals that grouping copynet can impressively improve the performance of PLAN BLEU-4 and PLAN ROUGE-L, and GAT can further improve the performance of PLAN ROUGE-L.}
  \label{tab:ablation}
  \resizebox{0.485\textwidth}{!}{
  \begin{tabular}{p{4cm} p{1cm} p{1cm} p{1cm}}
    \toprule
    Method & BLEU-4 & PB-4 & PR-L \\
    \midrule
     GGP + BART & \textbf{4.3} & \textbf{20.8} & \textbf{68.7} \\
     \midrule
     w/o graph networks & 4.2 & 20.8 & 67.9 \\
     w/o grouping copynet & 4.0 & 16.7 & 64.8 \\
    \bottomrule
  \end{tabular}
  }
\end{table}

Table \ref{tab:results} reveals our experimental results. Our model outperforms the baselines in terms of BLEU-4, PLAN BLEU-4 and PLAN ROUGE-L on three datasets, which indicates that our proposed method can better make the plan according to the given key phrase list without missing important input items in a long text. The most competitive baseline on ATG is PHVM, but BART performs a better result than PHVM. With GGP, it outperforms PHVM by 1.4\% on BLEU-4, 7.0\% on PLAN BLEU-4 and 4.0\% on PLAN ROUGE-L, indicating the effectiveness of our planner. As for Now You're Cooking and TaoDesc, DualEnc achieves state-of-the-art results on WebNLG~\cite{castro-ferreira-etal-2018-enriching}. However, it performs even worse than Random Planner on PLAN ROUGE-L on Now You're Cooking and much worse than BART on all metrics as it only considers instance-level information of a sentence. However, our GGP considers constructing plans from the current instance and the corpus perspective. On Now You're Cooking, our GGP outperforms DualEnc by 6.1\% on BLEU-4, 7.5\% on PLAN BLEU-4 and 18.4\% on PLAN ROUGE-L. And on TaoDesc, our proposed GGP outperforms DualEnc by 3.6\% on BLEU-4, 3.9\% on PLAN BLEU-4 and 2.1\% on PLAN ROUGE-L. In conclusion, GGP significantly outperforms baselines on PB-4 and PR-L on all of these three long text generation datasets, which proves that GGP can better control the long text generation process according to the given plan.

\subsection{Ablation Study}

To further investigate the effectiveness of graph networks and grouping copynet, we conduct ablation experiments on GGP. Table \ref{tab:ablation} reveals that graph networks can improve PLAN ROUGE-L by 0.8\%. And with grouping copynet, PLAN BLEU-4 is improved by 4.1\% while PLAN ROUGE-L is improved by 3.9\%, and there is almost no effect on BLEU-4 which is exploited to measure generated texts. It can be analyzed from the results that grouping copynet can learn plans well from the whole corpus and graph information can slightly help improve the results. The reason is that copynet can directly extract phrases from source data which are also revealed in generated plans. However, the model is more flexible with the graph structure as we can use the knowledge graph or the graph extracted from other corpora in our future work.

\subsection{Case Study}

\begin{figure}[t]
  \centering
  \includegraphics[width=0.6\linewidth]{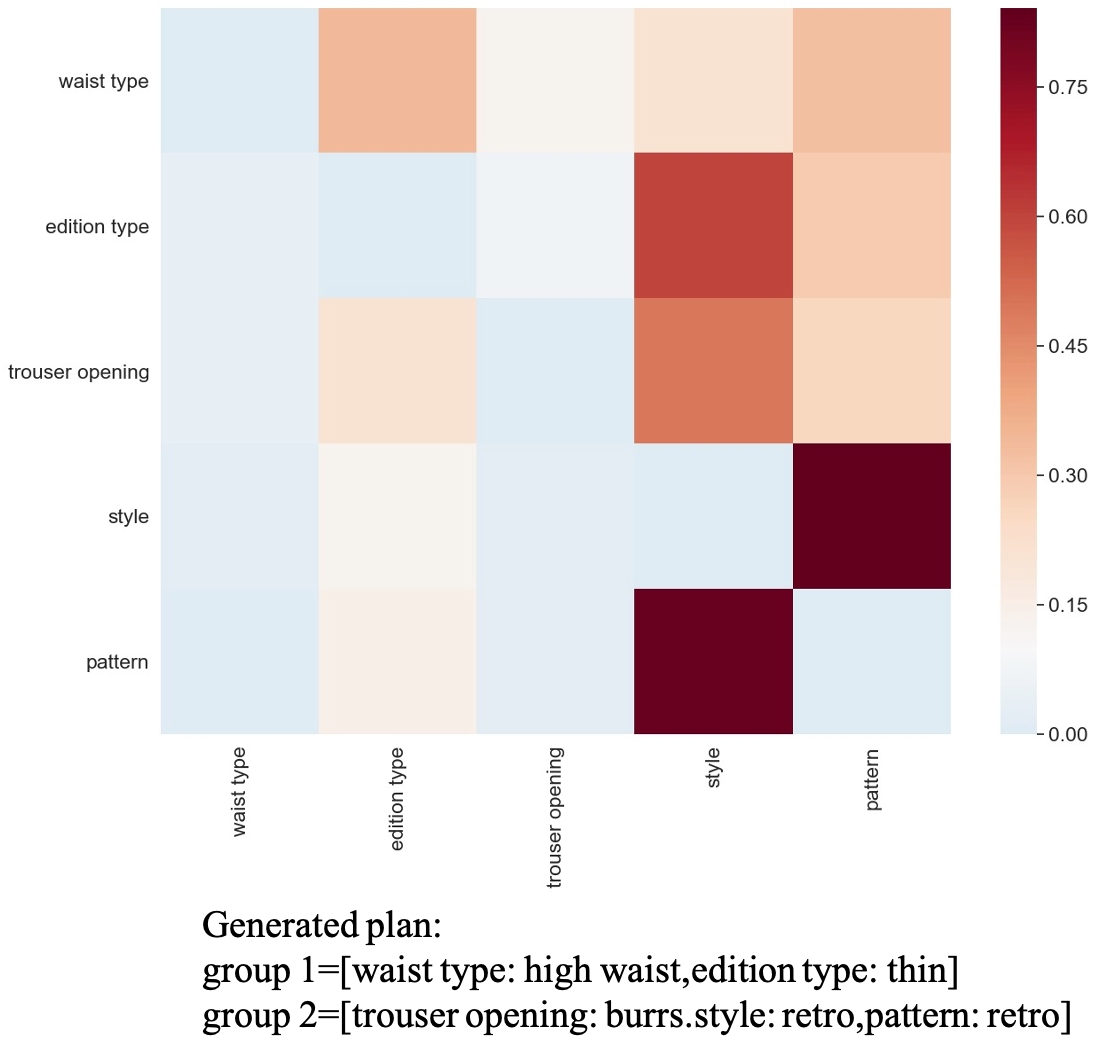}
  \caption{An example of graph attention and the corresponding generated plan which is exactly the same as the golden plan from the ATG dataset.}
  \label{fig:att_example}
\end{figure}

To observe how graph-based representations affect the planning process, we extract the graph attention from GAT and the corresponding generated plan from the ATG dataset. Figure~\ref{fig:att_example} reveals that attention scores in the same row are relatively high from "waist type" to "edition type", from "trouser opening" to "style" and from "style" to "pattern", which is how the first group and the second group are generated and both of them are combined into the final plan.

\section{Conclusion and Future Work}
In this work, we present a graph-based grouping planner~(GGP) for sentence planning and content planning together with graph-based information to explicitly control the process of long text generation. GGP combines the grouping copynet with graph neural networks to better capture the global information from the whole corpus and it can regroup plans from sentence-level to paragraph level. Experiments on three data-to-text corpora reveal that our model is more competitive to extract plans than state-of-the-art baselines. In the future, we will conduct more experiments on plan generation with the knowledge graph or graphs constructed from human live streaming corpus, and further apply it to our industrial scenarios.

\bibliographystyle{ACM-Reference-Format}
\bibliography{sample-base}

\end{document}